# EFFICIENT LICENSE PLATE RECOGNITION VIA PSEUDO-LABELED SUPERVISION WITH GROUNDING DINO AND YOLOV8

*Zahra E. Vargoorani*[1], *Amir Mohammad Ghoreyshi*[2], *Ching Y. Suen*[1]

[1]Concordia University, Montreal, Quebec, Canada
[2]Shahid Beheshti University, Tehran, Iran

## ABSTRACT

Developing a highly accurate automatic license plate recognition system (ALPR) is challenging due to environmental factors such as lighting, rain, and dust. Additional difficulties include high vehicle speeds, varying camera angles, and low-quality or low-resolution images. ALPR is vital in traffic control, parking, vehicle tracking, toll collection, and law enforcement applications. This paper proposes a deep learning strategy using YOLOv8 for license plate detection and recognition tasks. This method seeks to enhance the performance of the model using datasets from Ontario, Quebec, California, and New York State. It achieved an impressive recall rate of 94% on the dataset from the Center for Pattern Recognition and Machine Intelligence (CENPARMI) and 91% on the UFPR-ALPR dataset. In addition, our method follows a semi-supervised learning framework, combining a small set of manually labeled data with pseudo-labels generated by Grounding DINO to train our detection model. Grounding DINO, a powerful vision-language model, automatically annotates many images with bounding boxes for license plates, thereby minimizing the reliance on labor-intensive manual labeling. By integrating human-verified and model-generated annotations, we can scale our dataset efficiently while maintaining label quality, which significantly enhances the training process and overall model performance. Furthermore, it reports character error rates for both datasets, providing additional insight into system performance.

## 1. INTRODUCTION

Automatic License Plate Recognition (ALPR) systems encounter challenges due to environmental factors like rain, dust, high speeds, varied angles, and poor image quality [1]. Despite these challenges, ALPR remains vital for traffic management, security, and law enforcement [2]. Although numerous techniques using image processing and machine learning have been proposed, achieving high accuracy remains a challenge. Recent advancements show that convolutional neural networks [3] can effectively extract complex visual patterns, which holds tremendous potential for improving the performance and reliability of ALPR systems.

Grounding DINO [4] is a vision-language object detection model that leverages pre-trained image and text encoders to detect objects based on natural language prompts. It enables open-set object detection by grounding textual descriptions in visual content, allowing for flexible and scalable annotation. In our work, we utilized Grounding DINO [4] to automatically generate pseudo-labels for unlabeled images, significantly reducing the reliance on manual annotation. This approach supports a semi-supervised learning framework, where model-generated annotations complement a smaller set of human-labeled data, enabling efficient training on a larger dataset.

YOLOv8 Nano is an efficient and lightweight object detection model, and a variant of the YOLO (You Only Look Once) [5] model. It divides the image into a grid and simultaneously predicts bounding boxes and class probabilities [6]. Following the anchor-free approach, YOLOv8 Nano maintains robust detection performance across object sizes while remaining computationally efficient. Each grid cell predicts bounding boxes and associated confidence scores, which are evaluated using Intersection over Union (IoU). Built for edge devices in real-time applications, YOLOv8 Nano is optimized for rapid inference and high accuracy.

The main contributions of this work are:

1. Employing Grounding DINO, a semi-supervised learning approach, to label an unlabeled dataset recently collected from CENPARMI.

2. Proposing a one-stage solution for license plate detection and character recognition using YOLOv8.

### 1.1. Datasets

**UFPR-ALPR Dataset:** The UFPR-ALPR dataset [7], created by the Federal University of Paraná, includes 4,500 Brazilian license plate images from vehicles such as motorcycles, cars, and public transport. The images were taken with three cameras: GoPro Hero4 Silver, Huawei P9 Lite, and iPhone 7 Plus. The splits in the data are 40% for training, 40% for testing, and 20% for validation. All images are taken during the daytime and from moving vehicles. The range from the camera to license plates is 1 meter to 10 meters. Approximately 30% of the images are affected by challenging conditions such as intense sunlight, shadows, or dust. Such variations are substantial in designing and testing license plate recognition systems that work satisfactorily across real-world environments. Figure 1 presents a few sample images taken from the UFPR-ALPR [7] dataset.

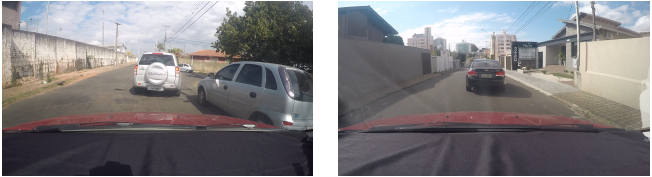

**Fig. 1**. Sample images from the UFPR-ALPR dataset

**CENPARMI Dataset:** The Centre for Pattern Recognition and Machine Intelligence (CENPARMI) has collected a dataset of 1,600 license plates in different parts of the United States and Canada. The images were captured under diverse lighting conditions, from various vehicle types—including automobiles, bicycles, and public transport—at distances ranging from 1 to 30 meters. Some images contain multiple license plates, making the task of identifying and recognizing each plate more difficult. This variety makes the set practical for real-world use and a worthwhile resource for building strong recognition systems. Figure 2 presents a few sample images taken from the CENPARMI dataset.

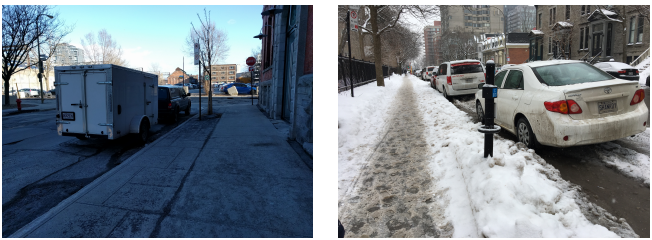

**Fig. 2**. Sample images from the CENPARMI dataset

Together, these datasets offer diverse and realistic scenarios for evaluating the performance and robustness of the proposed ALPR system.

## 2. RELATED WORK

Various approaches have been proposed for license plate recognition, ranging from traditional machine learning methods to deep learning-based systems.

Li [8] presented a method for overcoming the challenges of license plate detection in low-resolution conditions. The study blended AOD-Net for image enhancement, YOLOv5 for the detection of plates, and LPRNet with CRNN+CTC for recognizing characters. The system achieved an overall recognition rate of 91% on the CCPD dataset, with improved performance in cases that include low-quality images.

Researchers in [9] developed the LPRNet model to enhance license plate recognition in difficult conditions. The proposed system achieved 79.0% recognition accuracy on the CCPD2019 dataset, illustrating the difficulties and potential solutions to license plate recognition in difficult conditions.

A study in [10] proposed a deep-learning-based system using Convolutional Neural Networks and Long-Short-Term Memory Neural Networks to localize and identify Pakistani non-standardized, non-uniform-sized license plates with different fonts and styles. The system could identify the plates with 93.53% accuracy on the PLPD dataset, which was effective in handling multiple variations in license plate formats.

In reference [11] researchers proposed a multi-stage, real-time deep learning framework for car detection and license plate reading. The framework integrates two object detectors, one image classifier, and one multi-object tracker for detecting car models and license plates. Leveraging the redundancy of Saudi license plates with Arabic and English characters, the framework maximizes the recognition rate while maintaining real-time inference latency. Validated in real environments, it achieved a recognition rate of 93.5% on video streams.

In reference [7], researchers proposed a cascaded deep-learning pipeline for automatic car license plate detection and recognition under severe lighting and occlusion. The system used a YOLOv4-based detector followed by a CRNN-based recognition subsystem. To achieve robustness, the authors used a preprocessing stage that involves histogram equalization and spatial transformation. Assessed on the UFPR-ALPR database, the approach achieved a general recognition rate of 92.8%, demonstrating its practical usability even with somewhat reduced performance under night-time lighting and heavy occlusion.

These findings underscore the importance of customizing neural network architectures with domain-specific strategies to enhance ALPR performance.

## 3. PROPOSED METHODOLOGY

Although this study builds upon established approaches in license plate detection and recognition, it advances them through the integration of strategic techniques and the introduction of novel applications aimed at enhancing overall system performance. The proposed methodology includes the following main components:

### 3.1. License plate detection and character recognition

We propose a novel approach that integrates Grounding DINO [4], a vision-language object detection model that enables prompt-based detection and automatic annotation. This allows us to generate high-quality pseudo-labels for unlabeled data, reducing the need for extensive manual labeling. This framework generates pseudo-labels from unlabeled data, reducing the reliance on manually annotated datasets. Using these pseudo-labels, we trained the YOLOv8 Nano model, a lightweight and efficient architecture suitable for real-time processing on low-resource devices.

YOLOv8 Nano employs a distilled CNN backbone for feature extraction. It improves modules such as Cross Stage Partial Fusion (C2f) to reduce computational complexity [12]. It introduces the Spatial Pyramid Pooling-Fast (SPPF) module for capturing multi-scale contextual information with minimal addition to the model size. The design uses 3×3 convolutional kernels with a stride of 1 as the primary operation for spatial feature extraction. It inserts 1×1 convolutions at suitable locations to change the number of feature channels, adjusting the depth of feature maps without changing their spatial resolution [13]. Furthermore, YOLOv8 Nano employs an anchor-free detection head, simplifying detection and improving adaptability for objects of various sizes and aspect ratios [14].

The YOLOv8 Nano model was first employed in our pipeline to detect and isolate license plates from images. Subsequently, the same model was adapted for character recognition, effectively identifying single characters on the detected plates. This illustrates the versatility of YOLOv8 Nano in addressing multiple tasks within the ALPR pipeline. It also highlights the promise of incorporating weakly supervised approaches, such as the use of Grounding DINO [4] for automatic pseudo-labeling, which reduces the reliance on extensively labeled datasets.

After finalizing the model architecture, we focused on training and optimization.

### 3.2. Model training and optimization

We trained the detection and recognition models for over 100 epochs to achieve optimal convergence while avoiding overfitting. This number of epochs was selected empirically to balance training duration with model performance. The stochastic gradient descent optimizer was set for a learning rate of 0.01 to ensure stable convergence and accurate parameter updates. A batch size of 64 was used to balance computational efficiency with the model's ability to generalize across regions of interest within license plate images.

To evaluate our model's performance, we compared it with OpenALPR [15], a widely used commercial license plate recognition system. Developed by OpenALPR Technology Inc, this system supports license plate identification across multiple countries, including Canada, Brazil, the United States, Japan, and China. Comparing the results of our proposed model with those of OpenALPR [15] highlights the advantages and improvements of our approach. While OpenALPR [15] offers a comprehensive solution, our model demonstrates superior performance in challenging conditions, as evidenced by improved recognition accuracy and reduced character error rates.

Our results consist of three distinct sets. The first reports the Average Precision (AP) for the detection task performed using YOLOv8. The second provides Recall and Character Error Rate (CER) values for the recognition task. The third includes qualitative insights, such as confusion matrices per dataset, which aid in assessing character readability. Based on these results, we recommend several improvements to automatic license plate recognition systems.

### 3.3. Data augmentation and synthetic data generation

To improve the robustness and generalization of our recognition model, we applied various data augmentation techniques, including rotation, perspective distortion, and color channel adjustments. These transformations increased the visual diversity of the training data, enabling the model to handle variations commonly encountered in real-world environments, such as skewed angles and lighting inconsistencies.

In addition to traditional augmentations, we generated synthetic data by compositing cropped character images onto various license plate backgrounds. This synthetic dataset allowed us to simulate multiple license plate styles and environmental conditions, further enriching the training corpus.

By exposing the model to augmented real data and carefully crafted synthetic examples, we enhanced its ability to recognize characters under challenging conditions, including partial occlusion, varying illumination, and diverse plate designs. This combined strategy significantly contributed to the recognition system's overall accuracy and robustness in practical deployment scenarios.

## 4. RESULTS AND DISCUSSION

For training our detection and recognition model, we utilized the NVIDIA Tesla T4 GPU, which has 16 GB of GDDR6 memory and 2,560 CUDA cores. The GPU, based on NVIDIA's Turing architecture, offers massive computational power and parallel processing capabilities, enabling efficient handling of large-scale data and accelerating the training and optimization of our models.

### 4.1. Evaluation metrics

To assess our system's performance comprehensively, we employed three key evaluation metrics: mean Average Precision (mAP), Recall, and Character Error Rate (CER).

Mean Average Precision (mAP) is used to evaluate our model's object detection performance. It measures precision across different recall levels, capturing both the accuracy and completeness of the detections. Higher mAP values indicate better localization and classification of license plates and characters.

Recall quantifies the model's ability to identify all relevant instances correctly. In our context, it reflects how effectively the model detects all license plates or characters in the dataset. A high recall means fewer missed detections.

Character Error Rate (CER) evaluates the accuracy of the character recognition stage. It is calculated as the ratio of insertions, deletions, and substitutions required to transform the predicted text into the ground truth. Lower CER values indicate better performance, making it particularly important for assessing OCR quality in ALPR systems.

Together, these metrics comprehensively evaluate the model's performance in detection and recognition tasks, covering precision, completeness, and textual accuracy.

### 4.2. Detection results

Our YOLOv8 model's detection accuracy was evaluated using AP50:95, a strict measure considering varying IoU thresholds. The model with the UFPR-ALPR [7] dataset registered an average precision of 66.96%, indicating sound detection capability under challenging and changing conditions. Notably, the model performed even better on the CENPARMI dataset with an average precision of 79.67%, reflecting strong robustness and generalization ability across different data domains. These results validate the effectiveness of the YOLOv8 architecture for automatic license plate detection tasks and confirm its suitability for real-world applications. The results of the detection model for two datasets are shown in Table 1 and Figure 3 and 4 presents a few sample of detection results for CENPARMI and UFPR-ALPR [7] datasets respectively.

| Dataset | AP50:95 |
|---|---|
| UFPR-ALPR | 66.96% |
| CENPARMI | 79.67% |

**Table 1**. AP50:95 scores for license plate detection using YOLOv8 on two datasets

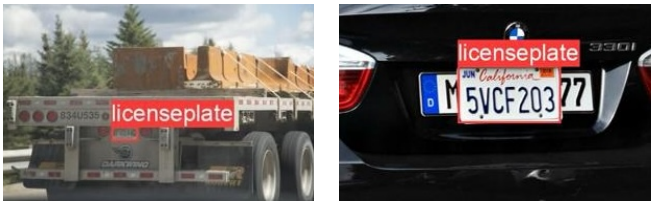

**Fig. 3**. Detection Sample Results of YOLOv8 Model on the CENPARMI Dataset

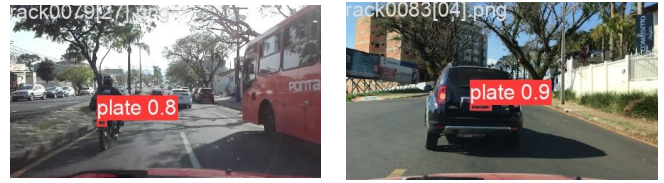

**Fig. 4**. Detection Sample Results of YOLOv8 Model on the UFPR-ALPR Dataset

### 4.3. Recognition results

Our proposed model was tested using two datasets: UFPR-ALPR [7] and CENPARMI. As shown in Table 2, the model achieved a Character Error Rate (CER) of 7.5% and a Recall of 91% on UFPR-ALPR [7]. It comfortably outperformed these results on the CENPARMI dataset, achieving a CER of 3.5% and a Recall of 94%. This improvement stems from the use of large-scale data augmentation and synthetic data generation during training, which enhanced the model's generalization to varied plate formats and conditions. The variability of the CENPARMI dataset—comprising license plates from different areas with different fonts, lighting conditions, and backgrounds—offered a robust test bed. These findings underscore the effectiveness of our training methodology and the generalization capacity of the YOLOv8-based recognition pipeline.

To further verify the robustness of our solution, we compared our method with two established recognition models: OpenALPR [15], a commercial off-the-shelf ALPR engine, and a CNN–RNN architecture trained with Connectionist Temporal Classification (CTC) Loss [16], referred to in [2]. As summarized in Table 2, the proposed model achieved superior Recall and lower character error rates across both datasets. On the UFPR-ALPR [7] dataset, our model reduced the CER from 15.5% (OpenALPR) to 7.5%, while increasing Recall from 55.8% to 91%. It also performed competitively compared to the CNN–RNN + CTC model [2], which achieved a CER of 5.3% and Recall of 92.1%. On the CENPARMI dataset, our proposed model achieved even more significant improvements, reducing the CER to 3.5% from 19.6% (OpenALPR) and 4.5% (CNN–RNN + CTC), while attaining the highest Recall of 94%.

Additionally, the YOLOv8-based model employs a lightweight architecture that enables fast inference on low-end devices such as smartphones, tablets, and embedded systems. This design ensures low computational overhead without compromising accuracy, making it suitable for real-time, on-device deployment in practical ALPR applications. Figures 5 and 6 present a few sample of character recognition results for CENPARMI and UFPR-ALPR [7] datasets respectively.

We also assessed the recognition model's performance on the CENPARMI and UFPR-ALPR [7] datasets along with quantitative metrics through confusion matrices. These matri-ces reveal the most common character-level confusions, which often occur due to visual similarities among certain characters. Such confusions are particularly probable in challenging scenarios, such as poor lighting, low resolution, or distortion.

| Model | Dataset | CER | Recall |
|---|---|---|---|
| OpenALPR | UFPR-ALPR | 15.5% | 55.8% |
| OpenALPR | CENPARMI | 19.6% | 80.2% |
| CNN–RNN + CTC | UFPR-ALPR | 5.3% | 92.1% |
| CNN–RNN + CTC | CENPARMI | 4.5% | 90.0% |
| YOLOv8 | UFPR-ALPR | 7.5% | 91% |
| YOLOv8 | CENPARMI | 3.5% | 94% |

**Table 2**. Comparison of CER and Recall across three models on UFPR-ALPR and CENPARMI datasets

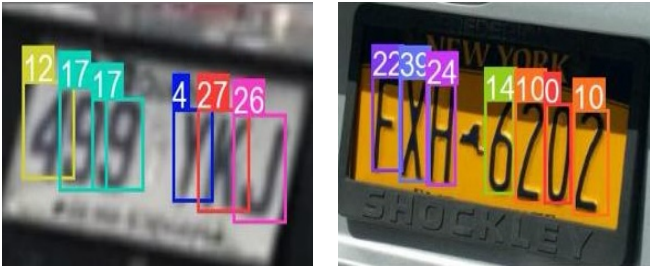

**Fig. 5**. Recognition Sample Results of YOLOv8 Model on the CENPARMI Dataset

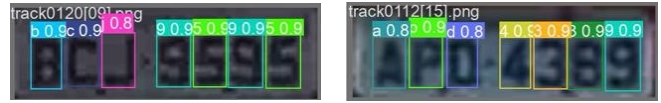

**Fig. 6**. Recognition Sample Results of YOLOv8 Model on the UFPR-ALPR Dataset

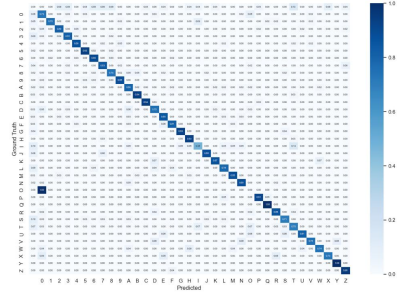

**Fig. 7**. Confusion Matrix for CENPARMI Dataset

Figure 7 demonstrates the confusion matrix on the test dataset of CENPARMI. The robustness of YOLOv8 is superb, but for a handful of character-level errors. For instance:

- **'O' and '0'**: These characters tend to look extremely similar due to their tightly rounded shapes.
- **'I' and 'T'**: The cross-bar on the letter 'T' is sometimes missing or faint, making it visually similar to 'I'.
- **'B' and '8'**: Their similar bowls and covered counters could lead to confusion, especially when the spacing is tight.

Figure 8 shows the confusion matrix of the UFPR-ALPR [7] dataset, where the number of recognition errors is slightly more significant. These are:

- Character similarities such as **'E' and '4'**, **'C' and '9'**, **'M' and 'W'**, and **'A' and 'W'** involve similar diagonal strokes, bowl shapes, or stem structures.
- There is additional confusion between **'I' and '1'**, **'J' and 'U'**, and **'D' and 'O'**, primarily because of smaller or stylistically similar fonts in registration plate layouts.

While such errors reflect natural visual ambiguity in character forms—driven partially by font design considerations like bowls, strokes, counters, and spurs—the YOLOv8 model still has great generalization and recognition performance on both data sets. The model's ability to process through these visually non-distinguishable cases with relatively low character error rates highlights its robustness and practical applicability for ALPR systems. These font-controlled results are also functional diagnostics for further tuning, such as tailored data enhancement or post-processing methods to reduce confusion between indistinct characters further.

## 5. CONCLUSION AND FUTURE WORK

This study presents a one-stage deep learning framework for automatic license plate detection and character recognition, inspired by YOLOv8. In addition, the Grounding DINO a vision-language object detection model that enables prompt-based detection and automatic annotation was employed to label a previously unlabeled dataset collected from CENPARMI, significantly reducing the need for manual annotation. The system's performance was evaluated on two datasets using character error rate, recall, and confusion matrices to assess character-level accuracy.

Future research directions include expanding the dataset, integrating advanced post-processing techniques, and optimizing the system for deployment on embedded or low-resource devices. Overall, this work contributes to the advancement of intelligent transportation systems by promoting more efficient, accurate, and scalable ALPR solutions for traffic monitoring and security applications.

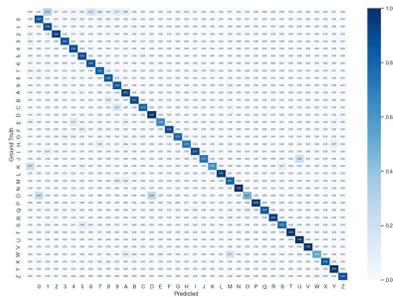

**Fig. 8**. Confusion Matrix for UFPR-ALPR Dataset